\documentclass[sigconf]{acmart}
\AtBeginDocument{%
  }

\usepackage{xcolor}
\usepackage{amsfonts}
\usepackage{listings}
\usepackage{color}
\usepackage{algorithm}
\usepackage{natbib}
\usepackage{twemojis}
\usepackage{tcolorbox}
\usepackage{mdframed}
\usepackage{xcolor}
\usepackage{mdframed}
\usepackage{graphicx}
\usepackage{algpseudocode} 
\usepackage{threeparttable}  
\usepackage{booktabs}   
\usepackage{subcaption}
\usepackage{tabularx}

\acmSubmissionID{488}

\copyrightyear{2026}
\acmYear{2026}
\setcopyright{cc}
\setcctype{by-nc-nd}
\acmConference[KDD '26]{Proceedings of the 32nd ACM SIGKDD Conference on Knowledge Discovery and Data Mining V.2}{August 09--13, 2026}{Jeju Island, Republic of Korea}
\acmBooktitle{Proceedings of the 32nd ACM SIGKDD Conference on Knowledge Discovery and Data Mining V.2 (KDD '26), August 09--13, 2026, Jeju Island, Republic of Korea}

\begin{document}
\title{Large Language Model-Powered Query-Driven Event Timeline Summarization in Industrial Search}

\author{Mingyue Wang}
\affiliation{
  \institution{Baidu Inc.}  
  \city{Beijing}
  \country{China}
}
\email{wangmy.mia@gmail.com}

\author{Xingyu Xie}
\affiliation{
  \institution{Baidu Inc.}
  \city{Beijing}
  \country{China}
}
\email{xiexingyu0506@gmail.com}

\author{Hang Yang}
\affiliation{
  \institution{Baidu Inc.}
  \city{Beijing}
  \country{China}
}
\email{yanghang.nlp@gmail.com}

\author{Li Gao}
\authornote{Corresponding author.}  
\affiliation{
  \institution{Baidu Inc.}
  \city{Beijing}
  \country{China}
}
\email{gaoli.sinh@gmail.com}

\author{Lixin Su}
\affiliation{
  \institution{Baidu Inc.}
  \city{Beijing}
  \country{China}
}
\email{sulixinict@gmail.com}

\author{Ge Chen}
\affiliation{
  \institution{Baidu Inc.}
  \city{Beijing}
  \country{China}
}
\email{chenge02@baidu.com}

\author{Dawei Yin}
\affiliation{
  \institution{Baidu Inc.}
  \city{Beijing}
  \country{China}
}
\email{yindawei@acm.org}

\author{Daiting Shi}
\affiliation{
  \institution{Baidu Inc.}
  \city{Beijing}
  \country{China}
}
\email{shidaiting01@baidu.com}

\renewcommand{\shortauthors}{Mingyue Wang et al.}


\begin{abstract}
Understanding how events evolve over time is essential for search engines handling queries about trending news. We present QDET (\textbf{Q}uery-\textbf{D}riven \textbf{E}vent \textbf{T}imeline Summarization), a production system deployed on Baidu Search that constructs focused event timelines to explain specific query events. Unlike traditional topic-centric approaches that aim for comprehensive coverage, QDET identifies and organizes sub-events closely relevant to the query from noisy candidate sets formed by millions of documents retrieved daily.
QDET incorporates two key innovations: (1) multi-task supervised fine-tuning with three auxiliary tasks—temporal ordering, causal judgment, and timeline completion—that enable compact models to match the performance of much larger general-purpose models in specialized domains; (2) reinforcement learning-based event concise summarization that enforces strict length constraints while maintaining semantic quality, achieving 88.2\% length compliance and outperforming 671B-scale models by 7.7 points in constraint satisfaction.
Our fine-tuned 7B parameter model achieves 76.2\% F1 score on timeline summarization, slightly surpassing the zero-shot performance of DeepSeek-R1-671B (76.1\% F1) while using only 1\% of its parameters—demonstrating that domain-specific optimization enables production-ready models with comparable quality at drastically reduced computational costs. Online A/B tests on Baidu Search validate real-world effectiveness, showing 5.5\% CTR improvement, 4.6\% longer dwell time, and 4.4\% deeper exploration compared to single-task baselines. We further demonstrate that timeline understanding transfers to heat prediction, confirming effective knowledge transfer to downstream tasks.
\end{abstract}





\maketitle

\section{Introduction}

Modern search engine users increasingly seek to understand evolv-
ing real-world events, especially breaking news and rapidly evolv-
ing trending topics. While traditional search engines return ranked document lists, users must still manually piece together
fragmented information across multiple sources to reconstruct the
underlying narrative. To address this challenge, large-scale search
systems such as Baidu are increasingly adopting Timeline Summarization (TLS)~\cite{tls_00,tls_02,tls_03,tls_04}, which condenses large document collections into
temporally structured event narratives that help users quickly com-
prehend event evolution.
Existing TLS research mainly falls into two categories~\cite{timeline_d1}. Event timeline summarization~\cite{els_0,els_1,els_2} focuses
on extracting evolving events from highly noisy text streams such
as tweets and organizing them into coherent temporal narratives.
Topic timeline summarization~\cite{topic_tls_1,topic_tls_3,topic_tls_4} instead aims to summarize
milestone developments associated with broader topics or entities
from document collections.

\begin{figure*}[htbp]
    \centering
    \begin{subfigure}[b]{0.48\textwidth}
        \includegraphics[width=\textwidth]{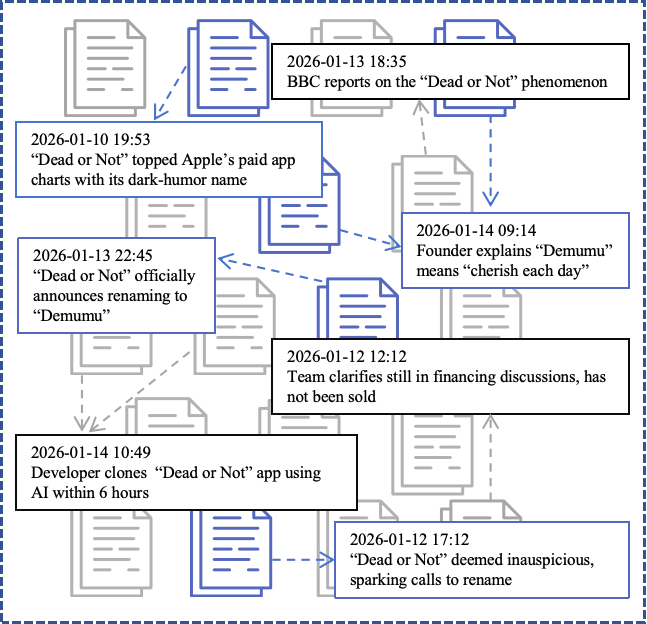}
        \caption{Retrieved candidate events from corpus}
        \label{fig:left}
    \end{subfigure}
    \hspace{0.03\textwidth} 
    \begin{subfigure}[b]{0.43\textwidth}
        \includegraphics[width=\textwidth]{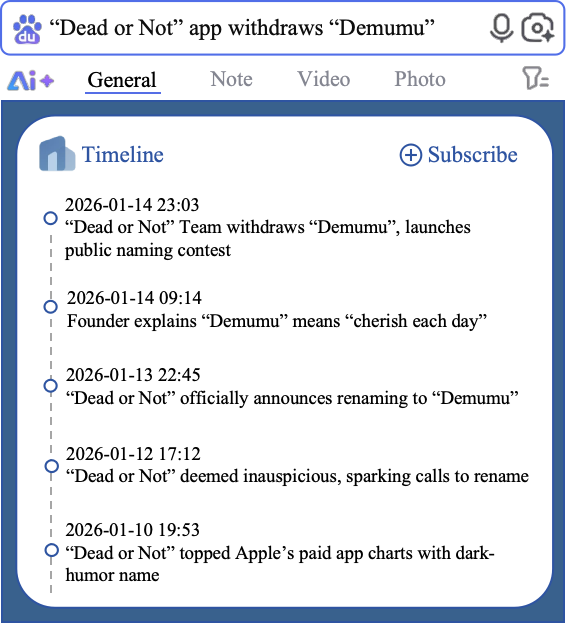}
        \caption{Event timeline summarization in Baidu Search}
        \label{fig:right}
    \end{subfigure}
    \caption{An example of timeline summarization from noisy candidate events. Given the query \emph{``Dead or Not app withdraws Demumu''}, the left subfigure illustrates the events retrieved from the corpus, while the right subfigure shows the real‑world application of event timeline summarization in Baidu Search.}
    \label{fig:wide_subfigures}
\end{figure*}

In real-world search scenarios, however, users often search for
a specific event and expect the system to explain how the event
emerged. We formulate this problem as query-driven event time-
line summarization, a search-oriented variant of event TLS. Query-
driven event TLS starts from a user-specified query event and aims
to reconstruct the key events and causal relationships leading to
its occurrence. Given a query event $e_q$, the system must retrieve
temporally and causally relevant events from noisy document col-
lections and organize them into a coherent timeline. As illustrated
in Figure~\ref{fig:wide_subfigures}, for the query event ``Dead or
Not app withdraws Demumu'', the system must identify 5 core
events from 20 candidate events collected over a 30-day retrieval
window.
This setting introduces several practical challenges. First, large-scale daily retrieval results contain substantial noise, requiring
accurate filtering of relevant events. Second, retrieved events must
be organized through temporal and causal reasoning to form a
coherent narrative. Third, generated event summaries must satisfy
strict display constraints imposed by industrial search interfaces.

Recently, large language models (LLMs)~\cite{llm0,llm1,llm2,llm3} have shown strong capabilities in timeline summarization and are increasingly being applied to TLS tasks. However, existing LLM-based timeline summarization approaches still face significant limitations in industrial search scenarios. Although large language models demonstrate strong summarization capabilities~\cite{timeline_d2,llm_tls_1,llm_tls_2}, they incur prohibitive serving costs for real-time deployment, lack effective mechanisms for query-driven temporal and causal reasoning, and struggle to satisfy strict output-length constraints required by industrial search interfaces. 

We argue that these limitations stem from treating query-driven event timeline generation as a summarization problem, while in practice it fundamentally requires structured reasoning over event relationships. Generating coherent timelines requires modeling temporal and causal relationships between events while preserving narrative completeness~\cite{temp_00,temp_02}. Motivated by this observation, we
propose \textbf{QDET} (\textbf{Q}uery-\textbf{D}riven \textbf{E}vent \textbf{T}imeline Summariza-
tion), a domain-specific multi-task learning framework designed
for industrial search deployment. QDET integrates three auxiliary
tasks—temporal ordering, causal judgment, and timeline comple-
tion—to provide inductive biases for query-driven relevance filter-
ing and narrative coherence. In addition, we introduce an reinforcement learning (RL)-based
concise event summarization framework that explicitly enforces
strict length constraints (5--15 characters) while preserving summary quality, achieving 88.2\% length compliance and outperforming
671B-parameter models by 7.7 percentage points.

Our fine-tuned 7B model achieves 76.2\% F1, slightly exceeding
DeepSeek-R1-671B’s 76.1\% zero-shot score while using only ap-
proximately 1\% of the parameters, demonstrating production-ready
quality at drastically lower cost. Since October 2025, QDET has
been deployed on Baidu Search, serving millions of daily trending-
event queries in real time, and extends timeline understanding to
heat prediction for adaptive visual emphasis within timelines.

The main contributions of this work include:
\begin{itemize}
\item We propose the QDET framework combining multi-task learning with RL-based constraint satisfaction, enabling compact models (7B parameters) to approach the performance of general-purpose models over 100$\times$ their size in specialized timeline generation tasks.
\item We formalize the query-driven event timeline generation problem and design three auxiliary tasks—temporal ordering, causal judgment, and timeline completion—which enhance the model’s capabilities in narrative filtering and coherence construction.
\item We demonstrate that structured timeline understanding transfers effectively to downstream heat prediction, providing the first investigation of intra-timeline engagement dynamics for query-driven event timelines in industrial search systems.
\end{itemize}

\section{Related Work}

\subsection{Timeline Summarization}

Recent work has leveraged LLMs to improve timeline summarization. Hu et al.~\cite{timeline_d1} proposed LLM-TLS, which incrementally updates timelines by matching incoming documents to existing clusters via vector retrieval. Wu et al.~\cite{timeline_d2} introduced an iterative self-questioning framework that progressively retrieves information through LLM-generated questions to generate comprehensive timelines. Other work has explored event detection~\cite{e_dection} and argument extraction~\cite{multi_extraction}, treating events as independent instances. While these methods demonstrate strong capabilities, they still operate in a topic-centric manner, aiming for broad coverage rather than focusing on query-driven explanation.

While these methods demonstrate strong timeline generation capabilities, they mainly focus on constructing globally coherent timelines for broad topics or evolving events. In contrast, search scenarios often require query-driven timelines that explain the temporal and causal context behind a specific query event. QDET addresses this gap by identifying and organizing events that help users understand how a query event emerged, rather than exhaustively summarizing all developments within a topic.

\subsection{Temporal and Causal Reasoning}

Generating query-driven event timelines requires understanding temporal relationships and causal dependencies between events. Recent work~\cite{xiong2024large,time-r1,li2021temporal,su2025enhancing,chen2026rag} has explored enhancing LLMs' temporal ordering capabilities through targeted training. Xiong et al.~\cite{xiong2024large} fine-tuned LLMs on text-to-temporal-graph translation using synthetic data, while Liu et al.~\cite{time-r1} introduced a progressive RL curriculum developing temporal intelligence through timestamp inference, event ordering, and future event prediction. For causal reasoning, Li et al.~\cite{li2021temporal} and Su et al.~\cite{su2025enhancing} demonstrated that leveraging LLM knowledge for event causality identification significantly improves story understanding tasks. These works show that LLMs can acquire temporal and causal reasoning capabilities through task-specific training.

However, existing timeline summarization methods rarely integrate temporal and causal reasoning into a unified generation framework. Prior work typically focuses on isolated subtasks such as temporal ordering or causal extraction, often operating on complete narratives rather than noisy candidate event sets retrieved from search corpora. QDET addresses this limitation by jointly training on three auxiliary tasks—temporal ordering, causal judgment, and timeline completion—which enhance the model’s ability to generate coherent query-driven timelines through reasoning over event chains rather than isolated event pairs.

\subsection{Engagement Prediction}
Understanding and predicting user engagement is crucial for content curation and information dissemination in social media. Recent work has begun applying LLMs to social media trend forecasting~\cite{chen2024predicting,Buzz,zheng2025fusing}. Chen et al.~\cite{Buzz} combined LLMs' contextual reasoning capabilities with traditional regression models' numerical precision for hashtag popularity prediction. Zheng et al.~\cite{zheng2025fusing} proposed TweetFusion-LLM, which reprograms topic popularity time series into LLaMA-2's~\cite{llama2} embedding space, fusing temporal dynamics with tweet semantics for forecasting.

Inspired by these works, we further explore heat prediction as a downstream application of timeline understanding. Unlike prior studies that predict absolute popularity metrics for isolated topics, we model relative intra-timeline heat patterns to support adaptive presentation within generated timelines. By fine-tuning our timeline summarization model on heat-labeled data, we demonstrate that structured timeline representations transfer effectively to heat prediction, enabling intelligent visual emphasis of salient events in search interfaces. To our knowledge, this is the first work investigating intra-timeline engagement dynamics for query-driven event timelines in industrial search systems.

\section{Methodology}
\label{sec:method}
\subsection{Problem Formulation and System Overview}
\label{sec:formulation}
We formally define the problem of query-driven event timeline summarization from streaming documents.

\textbf{Document Stream.}
A real-time document stream is defined as a temporally ordered sequence $\mathcal{D} = \{d_1, d_2, \ldots, d_n\}$, where each document $d_i = (c_i, t_i)$ consists of textual content $c_i$ and publication timestamp $t_i$. Documents arrive in chronological order, i.e., $t_i \leq t_j$ whenever $i \leq j$.

\textbf{Event.}
An event is defined as a tuple $e = (s_e, t_e, d_e, D_e)$,
where $s_e$ denotes the event summary, $t_e$ denotes the timestamp of the representative source
document $d_e$ and is used as the canonical timestamp of the event,  and $D_e$ denotes the set of supporting documents associated with the event. In practice, an event corresponds to a cluster of multi-source documents reporting the same real-world event.

\textbf{Query Event.}
A query event $e_q$ denotes a target event that users
want to understand, typically corresponding to a trending event
recommended by the search platform. The goal of the system is to provide events that are temporally and causally relevant to $e_q$, in order to help users understand the development process.

\textbf{Query-Driven Event Timeline.}
Given a query event $e_q$, the task is to generate an event timeline $\mathcal{T} = \langle e_1, e_2, \ldots, e_m \rangle$ that explains the background, progression,
and related developments of the query event $e_q$. The timeline must satisfy:
\begin{itemize}
    \item \textbf{Temporal ordering:} $t_{e_1} \leq t_{e_2} \leq \cdots \leq t_{e_m}$
    \item \textbf{Query-driven:} $e_q \in \mathcal{T}$ serves as the focal point, typically near the end. Each $e_i \in \mathcal{T}$ is semantically or causally related to $e_q$, forming a coherent narrative around the query event.
\end{itemize}

This formulation distinguishes our task from traditional timeline summarization, which typically generate topic-centric timelines covering all major developments. In contrast, query-driven event timelines are focused and explanatory, selecting only events that contribute to understanding the specific query event. We summarize the key notations in Table~\ref{tab:notations}.

\begin{figure*}[!htbp]
\centering
\includegraphics[width=0.99\textwidth]{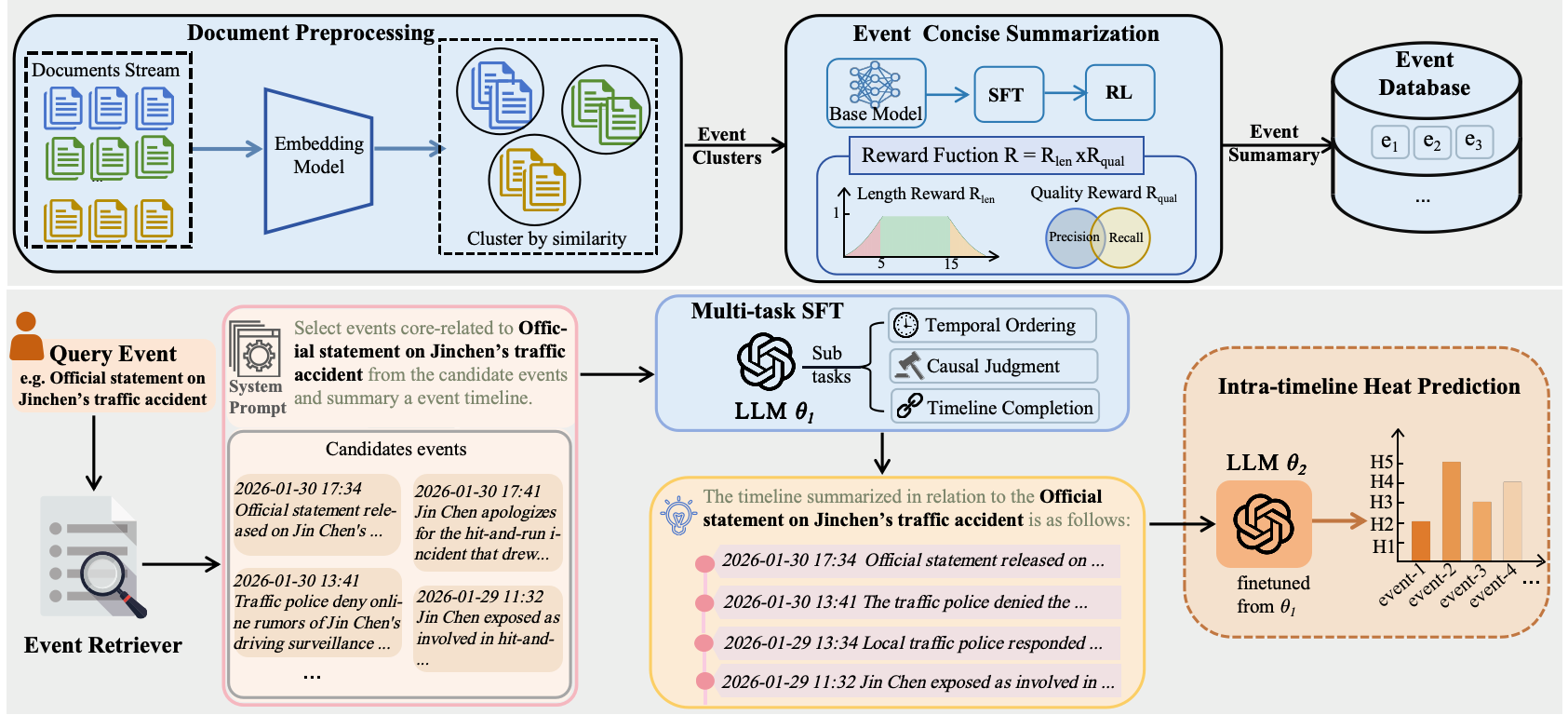}
\caption{System Architecture of QDET: A Query-Driven Event Timeline Summarization Framework consisting of four stages: (1) Document Preprocessing, which transforms real-time document streams into event clusters; (2) RL-Based Event Concise Summarization, which generates a concise summary for each event cluster and stores it in an event database; (3) Multi-Task Fine-Tuned LLM for Timeline Summarization, which, given a specific query, retrieves relevant candidate events from the database and generates a coherent, explanatory event timeline; and (4) Intra-timeline Heat Prediction, an auxiliary module designed to estimate the relative heat level of each event within timeline.}
\label{fig:architecture}
\end{figure*}

We propose QDET, a query-driven event timeline summarization framework deployed in Baidu Search (as shown in Figure~\ref{fig:architecture}). The framework first preprocesses the raw document stream $\mathcal{D}$ into structured event representations using lightweight models, clustering documents referring to the same real-world event based on similarity into event clusters. Then, it employs a RL-based method to generate concise, presentation-ready event summaries for each event cluster that meet the display requirements of the search platform. Finally, given a query event $e_q$, it retrieves candidate events and leverages a multi-task supervised fine-tuned LLM to produce a coherent and explanatory event timeline $\mathcal{T}$. The core contribution of this work lies in this LLM-based timeline summarization stage.

\begin{table}[t]
\centering
\caption{Summary of Key Notations}
\label{tab:notations}
\small
\begin{tabular}{cl}
\toprule
\textbf{Symbol} & \textbf{Description} \\
\midrule
$\mathcal{D} = \{d_1, d_2, \ldots, d_n\}$ & Real-time document stream \\
$d_i = (c_i, t_i)$ & Document with content and timestamp \\
$e = (s_e, t_e, d_e, D_e)$ & Event representation \\
$s_e$ & Event summary \\
$d_e$ & Representative source document of the event \\
$D_e$ & Multi-source document set of the event \\
$e_q$ & Query event  \\
$\mathcal{T} = \langle e_1, \ldots, e_m \rangle$ & Query-driven event timeline \\
\bottomrule
\end{tabular}
\end{table}

\subsection{Document Preprocessing}
\label{sec:preprocess}

In Baidu Search, the document stream $\mathcal{D}$ arrives at a scale of millions of documents per day. Given the throughput requirements of large-scale real-time ingestion, directly applying LLMs is computationally impractical. Therefore, we employ a lightweight dense text encoder
that produces 1024-dimensional embeddings for efficient event
clustering, similar in design to dense retrieval models such as BGE~\cite{bge} and E5~\cite{e5}. Detailed training settings are described in Section~\ref{sec:imple_detail}.

Documents reporting the same real-world event are grouped into event clusters through embedding-based similarity matching within
a configurable sliding time window (30 days in our deployment).
Incoming documents are incrementally assigned to existing event clusters based on embedding similarity within the temporal window; otherwise, a new event cluster is created.
To support scalable clustering over continuously updated embedding
indexes, we employ approximate nearest neighbor (ANN) search~\cite{ann}.
After generating concise summaries for each event cluster through the event concise summarization stage (Section~\ref{sec:namegen}), the resulting summarized events are stored in the event database and used for downstream timeline retrieval and summarization.

\subsection{Event Concise Summarization}
\label{sec:namegen}
In Baidu Search, our objective is to generate concise event summaries
for each event cluster in order to efficiently convey core event information to users. Given an event cluster $D_e$ produced by the upstream
clustering stage, we first select a representative source document $d_e$
from the cluster (typically the earliest document in time) that best
reflects the overall event. Based on the title and content of $d_e$, we then generate a corresponding event summary $s_e$.

We believe that a well-crafted event summary should accurately
capture the core semantics of the event within 5 to 15 Chinese
characters. This length constraint is empirically derived from our
analysis of Baidu’s production data: statistics from millions of
news events show that over 90\% of events can be accurately summarized within 15 characters, while summaries shorter than 5
characters often omit critical information. In addition, the quality
and writing style of original news titles vary significantly across
sources, making a unified summarization process necessary for
consistent event representation.

To generate summaries satisfying these constraints, we adopt a
two-stage training framework before events proceed to downstream
retrieval and timeline summarization. First, we perform supervised
fine-tuning (SFT) on a compact generative language model to establish semantic summarization capability. We then apply RL to optimize a reward function balancing summary
conciseness and semantic quality.

\textbf{Reward Function Design.}
Let the generated summary be $s_e$ and the reference summary be $s_{\text{ref}}$. We design a composite reward balancing length compliance with semantic quality:
\begin{equation}
R(s_e; s_{\text{ref}}) = R_{\text{len}}(s_e) \cdot R_{\text{qual}}(s_e, s_{\text{ref}})
\label{eq:reward}
\end{equation}

For length compliance, we employ an asymmetric Gaussian penalty that more strongly penalizes under-length summaries (which lose critical information) than over-length summaries (which merely reduce conciseness):
\begin{equation}
R_{\text{len}}(s_e) = 
\begin{cases}
\exp\left(-\lambda_s (l_{\min} - |s_e|)^2\right) & \text{if } |s_e| < l_{\min} \\
1 & \text{if } l_{\min} \leq |s_e| \leq l_{\max} \\
\exp\left(-\lambda_l (|s_e| - l_{\max})^2\right) & \text{if } |s_e| > l_{\max}
\end{cases}
\label{eq:length_reward}
\end{equation}
where $[l_{\min}, l_{\max}] = [5, 15]$, $\lambda_s = 0.5$, and $\lambda_l = 0.3$ denote the penalty coefficients for under-length and over-length summaries, respectively. The asymmetry reflects our observation that users can tolerate slight verbosity but not missing critical facts.

For semantic quality, we compute a weighted F-score that emphasizes recall over precision:
\begin{equation}
R_{\text{qual}}(s_e, s_{\text{ref}}) = \frac{(1 + \alpha^2) \cdot P \cdot R}{\alpha^2 \cdot P + R}
\label{eq:qual_reward}
\end{equation}
where $P$ and $R$ denote character-level precision and recall, and $\alpha = 1.5$ weights recall more heavily. This choice reflects the priority of information preservation: it is more important to retain key information than to avoid extraneous characters in the generated summary.

An important design consideration is that the generated event
summaries are directly used as the final textual units in the output
timeline. Downstream timeline summarization modules therefore
focus on selecting and ordering existing event summaries rather
than rewriting them, which substantially reduces generation complexity and improves consistency across the system. After summary
generation, each summarized event is stored in the event database
together with its associated source document ID for downstream
retrieval and timeline summarization.
In the deployed Baidu Search system, users can directly navigate
from a timeline event to its linked full-text source document, which
includes metadata such as publication source, author, and timestamp. Therefore, event summaries serve as concise navigational
anchors rather than standalone authoritative descriptions, ensuring
transparency and traceability to the original content.

\subsection{Event Timeline Summarization}
\label{sec:event-timeline}

Given a query event $e_q$, this stage retrieves candidate events and generates a coherent event timeline $\mathcal{T}$ providing contextual and temporal understanding of $e_q$. This is the core contribution of QDET.

\subsubsection{Event Retriever}
\label{sec:event-retriver}
We encode the query event using a separate 256-dimensional domain-adapted retriever encoder, obtaining the query embedding $\mathbf{x}_q \in \mathbb{R}^{256}$. Event embeddings are computed from the concise event summaries generated in Section~\ref{sec:namegen}. Detailed implementation and training settings are provided in Section~\ref{sec:imple_detail}.
Candidate events are retrieved from the event database within a sliding time window \([t_{e_q} - \Delta t, t_{e_q}]\), where \(\Delta t = 30\) days. This window size is empirically determined on the validation set, where we observe that most rapidly evolving news events in real-world hot news scenarios experience most of their major developments within 30 days. Larger windows introduce substantially more noise while providing limited additional contextual benefit. While this fixed-window design is effective for the majority of news scenarios, it may be less suitable for long-running events such as legal cases or prolonged social incidents.

Within this time window, we compute cosine similarity between $\mathbf{x}_q$ and all event embeddings using ANN index, and select the top-$k$ most similar events to form the retrieval set $\mathcal{R}$. Based on a recall-cost trade-off analysis, we set $k=20$.

It is important to note that semantic similarity does not necessarily imply relevance to explaining the query event; retrieved candidates typically contain: (1) core events that directly contribute to understanding $e_q$; (2) peripheral events belonging to the same broad topic but lacking direct contribution to the narrative; (3) redundant reports describing the same real-world incident from different sources; and (4) semantic noise that is topically similar but narratively irrelevant. We intentionally avoid hard threshold filtering at this stage to maintain high recall, delegating the precision-critical filtering to the LLM-based summarization stage, which is better equipped to reason about event relevance and narrative coherence.

\subsubsection{Multi-task Supervised Fine-Tuning}
To address the limitations of general-purpose LLMs in distinguishing between direct and thematic relevance among events, we employ a multi-task fine-tuning approach. This method combines the main timeline summarization task with a set of targeted auxiliary tasks to systematically enhance the model’s relevance reasoning capabilities, particularly in temporal and causal reasoning. The training data consists of: (1) a main timeline summarization dataset, and (2) three auxiliary task datasets focusing on temporal ordering, causal judgment, and timeline completion, respectively. These auxiliary tasks are designed to build the foundational capabilities required by the main task. Their training data is mixed with that of the main task, shuffled, and used for joint supervised fine-tuning.

\textbf{Main Task (Timeline Summarization):} Given a query event $e_q$ and a set of candidate events $\mathcal{R}$ (typically containing $m=20$ candidates), the model generates an explanatory event timeline:
\begin{equation}
f_{\text{timeline}}: (e_q, \mathcal{R}) \rightarrow \mathcal{T} = \langle e_1, \ldots, e_{n-1}, e_n \rangle, \quad n \ll m
\end{equation}
The output $\mathcal{T}$ is a filtered and ordered event sequence forming a coherent timeline that explains how $e_q$ occurred. This requires the model to simultaneously filter noise from $\mathcal{R}$, identify relevant events, and arrange them into a temporally and narratively coherent sequence.

\textbf{Auxiliary Task:} In order to further enhance the LLM's ability to reason over event relevance and temporal structure in the context of query-driven timeline summarization, we have designed three auxiliary tasks. 1) temporal ordering: given a set of events presented in shuffled order along with their timestamps, the model is tasked with reconstructing their chronological sequence. This training objective establishes the model's foundational event ordering capability. 2) causal judgment: given a query event $e_q$ and a candidate event $e_c$, the model classifies their relationship as $y \in \{\textit{cause}, \textit{result}, \textit{irrelevant}\}$, where \textit{cause} indicates $e_c$ contributes to $e_q$, \textit{result} indicates $e_c$ is a consequence of $e_q$, and \textit{irrelevant} indicates no direct relationship. This task trains the model to reason over fine-grained event relationships, a capability essential for generating coherent timelines. 3) timeline completion: given a partial timeline $\langle e_1, \ldots, \texttt{[MASK]}, \ldots, e_q \rangle$ with a missing intermediate event, the model infers the missing element through multi-hop reasoning. This builds the narrative coherence understanding essential for generating complete explanatory timelines. The details of the prompt design for each task are shown in Table~\ref{tab:task_prompts}.

\begin{table}[t]
\centering
\caption{Prompt design examples for auxiliary tasks.}
\label{tab:task_prompts}
\small
\begin{tabular}{l|p{0.6\linewidth}}
\toprule
\textbf{Task} & \textbf{Prompt} \\
\midrule
Temporal ordering &
Sort the following events in chronological order. \newline
Events: \{$e_1, e_2, \ldots, e_n$\} \newline
System: $\langle e_{\pi(1)}, e_{\pi(2)}, \ldots, e_{\pi(n)} \rangle$ \\
\midrule
Causal judgment &
Classify the relationship between the query event and the candidate event as cause, result, or irrelevant. \newline
Query event: \{$e_q$\} \newline
Candidate event: \{$e_c$\} \newline
System: \{cause / result / irrelevant\} \\
\midrule
Timeline completion &
Infer the missing event that completes the event timeline leading to the query event. \newline
Timeline: $\langle e_1, \ldots,$ \texttt{[MASK]}$, \ldots, e_q \rangle$ \newline
System: \{$e_k$\} \\
\bottomrule
\end{tabular}
\end{table}

\textbf{Supervised Fine-Tuning:} The training objective is to maximize the conditional probability of the target output given the task-specific prompt. Let $x$ denote the prompt and $y$ denote the target output, assuming $p(y|x) = \prod_{i=1}^{|y|} p(y_i | y_{0:i-1}, x)$, the training objective is to minimize the negative log-likelihood:
\begin{equation}
\mathcal{L}_{\text{SFT}}(\theta) = -\mathbb{E}_{(x, y) \sim \mathcal{D}_{\text{mix}}} \sum_{i=1}^{|y|} \log \pi(y_i | y_{0:i-1}, x; \theta)
\label{eq:sft}
\end{equation}
where $\mathcal{D}_{\text{mix}}$ denotes the mixed training data consisting of the main timeline summarization dataset and all auxiliary task datasets, $\pi(\cdot)$ and $\theta$ denote the timeline summarization model and its parameters, respectively. Following standard practice, the loss is computed only over the target output tokens and excludes the prompt tokens.

\section{Experiments}
\label{sec:experiments}
\subsection{Datasets}

Our experimental data originates from Baidu's hot event operation platform. We collected 8,000 human-annotated event timelines spanning July 2023 to July 2025, each representing a focused narrative leading to a specific query event. These timelines are summarized by professional editors who identify milestone events and organize them into coherent storylines based on user query intent, with each timeline undergoing rigorous verification of event relevance, temporal ordering, and narrative coherence.

We allocate 6,000 timelines for training, 800 for validation, and 1,200 for offline testing. The three splits exhibit consistent statistical distributions: training timelines average 7.5 events (std=2.3), validation timelines average 7.6 events (std=2.4), and test timelines average 7.8 events (std=2.5). Domain distributions are also balanced across splits, with politics (28--30\%), finance (21--23\%), entertainment (18--20\%), technology (17--19\%), and others (12--14\%) maintaining similar proportions in each partition. This consistency ensures that validation performance reliably predicts test performance and that our models generalize across the full domain spectrum.

From the 6,000 training timelines, we generate the multi-task training dataset: the 6,000 timelines directly serve as training data for event timeline summarization (main task); for temporal ordering, we randomly shuffle the event order in each timeline to generate 6,000 training samples; for causal judgment, we extract query-candidate event pairs from the timelines to construct 3,500 annotated training samples; for timeline completion, we randomly mask 1--2 intermediate events in the timelines to generate 3,500 training samples. 

For event concise summarization, we additionally collect 50K event clusters from the same platform's event database spanning 2023--2025, with each cluster accompanied by human-annotated reference summaries (concise texts of 5--15 characters capturing core event semantics). We use 45K clusters for RL training and hold out 5K for evaluation.

\subsection{Evaluation}

Our evaluation quantifies two core requirements of timeline summarization: accurate event selection and correct chronological ordering. For offline evaluation, we measure event selection quality using standard Precision, Recall, and F1 metrics; timeline content fidelity via Alignment-based ROUGE (AR-1) \cite{ar1} — a metric that evaluates the textual overlap between the summarized timeline and the reference timeline, with its alignment being based on temporal and semantic distance. Temporal ordering accuracy is valued through Kendall’s Tau ($\tau$) \cite{dehling2017testing}, where values closer to 1 indicate stronger sequence agreement. For event concise summarization, we evaluate semantic overlap with ROUGE-1/2/L and practical utility via length compliance, defined as the percentage of generated summaries falling within the 5–15 character range.

To gauge real-world utility, we conduct live A/B tests on Baidu Search, tracking key engagement metrics: click-through rate (CTR), dwell time on timeline pages, and exploration depth (average event clicks per timeline). These reflect whether users find the timelines relevant and engaging enough to explore further.

\subsection{Implementation Details}
\label{sec:imple_detail}
Unless stated otherwise, all LLM models for event timeline summarization are trained for 10 epochs with learning rate $2 \times 10^{-5}$, batch size 16, using AdamW optimizer with 10\% linear warmup and cosine learning rate decay. All experiments are run on 8 NVIDIA A100 (80GB) GPUs.

For event concise summarization (Section~\ref{sec:namegen}), we use Qwen2.5-7B-Instruct as the base model and adopt a two-stage training pipeline. 
We first perform supervised fine-tuning on 45K event-summary pairs.
For Proximal Policy Optimization (PPO) training, we randomly sample 20K clusters from the training
set due to the substantially higher computational cost of RL training.
PPO training uses a clipping threshold of 0.2 and a KL-divergence penalty coefficient of 0.05.
The reward function hyperparameters are set as: $\alpha = 1.5$, $\lambda_s = 0.5$, $\lambda_l = 0.3$.

We further employ two independently trained embedding models for document clustering and event retrieval, respectively. For document clustering (Section~\ref{sec:preprocess}), we use a 1024-dimensional encoder trained with contrastive learning (InfoNCE~\cite{infonce}) on approximately 1M document-document pairs constructed from event clusters, where documents from the same cluster are treated as positive pairs and documents from different clusters as negatives. The higher-dimensional representation preserves fine-grained semantic information necessary for accurate event-level clustering.
For event retrieval (Section~\ref{sec:event-retriver}), we use a separate 256-dimensional encoder, where event embeddings are computed from the concise event summaries generated in Section~\ref{sec:namegen}. The retriever is trained on approximately 100K event-level positive/negative pairs derived from manually normalized event matching annotations using the same contrastive learning objective. The lower-dimensional representation improves retrieval efficiency and reduces memory consumption while maintaining sufficient retrieval quality for downstream timeline summarization.

\subsection{Offline Experiments}

\subsubsection{Results of Different LLMs}

We evaluate the effectiveness of our multi-task SFT method by comparing models at different scales and training strategies. Table~\ref{tab:main_results} presents results for three configurations: (1) zero-shot prompting without task-specific fine-tuning, (2) 5-shot prompting with in-context demonstrations, and (3) models fine-tuned using our proposed multi-task learning framework.

\begin{table}[t]
\centering
\caption{Event timeline summarization performance across model scales and training strategies. Entries marked $\dagger$ are evaluated without task-specific fine-tuning using either zero-shot or 5-shot prompting. Models below the horizontal line are fine-tuned with the proposed multi-task learning framework.}
\label{tab:main_results}
\resizebox{\linewidth}{!}{
\begin{tabular}{lccccc}
\toprule
Base Model & P$\uparrow$ & R$\uparrow$ & F1$\uparrow$ & AR-1$\uparrow$ & $\tau\uparrow$ \\
\midrule
Qwen2.5-3B-Instruct (0-shot)$^\dagger$ & 59.2 & 60.1 & 59.6 & 59.3 & 0.65 \\
Qwen2.5-3B-Instruct (5-shot)$^\dagger$ & 60.5 & 61.8 & 61.1 & 60.5 & 0.67 \\
Qwen2.5-7B-Instruct (0-shot)$^\dagger$ & 62.7 & 64.8 & 63.7 & 63.1 & 0.68 \\
Qwen2.5-7B-Instruct (5-shot)$^\dagger$ & 64.1 & 66.2 & 65.1 & 64.5 & 0.70 \\
DeepSeek-R1-671B (0-shot)$^\dagger$ & \underline{73.6} & \underline{78.7} & \underline{76.1} & \underline{74.5} & \underline{0.89} \\
DeepSeek-R1-671B (5-shot)$^\dagger$ & 73.9 & 79.2 & 76.4 & 74.8 & 0.89 \\
\midrule
Qwen2.5-3B-Instruct (fine-tuned) & 66.8 & 74.5 & 70.4 & 68.1 & 0.83 \\
Qwen2.5-7B-Instruct (fine-tuned) & \textbf{73.2} & \textbf{79.5} & \textbf{76.2} & \textbf{73.8} & \textbf{0.87} \\
\bottomrule
\end{tabular}
}
\end{table}

Among the fine-tuned models, Qwen2.5-7B-Instruct achieves the best overall performance, reaching 76.2\% F1 and 73.8\% AR-1. Despite containing only 7B parameters (approximately 1\% of the 671B model size), its performance is comparable to DeepSeek-R1-671B under both zero-shot (76.1\% F1) and 5-shot (76.4\% F1) settings. This demonstrates that task-specific multi-task fine-tuning enables compact models to approach the performance of general-purpose models that are over two orders of magnitude larger in specialized domains.
Although 5-shot prompting consistently improves the performance of all base models over their zero-shot counterparts, the gains remain substantially smaller than those obtained through supervised fine-tuning. For example, Qwen2.5-7B-Instruct improves from 63.7\% to 65.1\% F1 under 5-shot prompting, while multi-task fine-tuning further boosts performance to 76.2\% F1. This result suggests that in-context demonstrations alone are insufficient for robust query-driven timeline reasoning under noisy retrieval settings.

\subsubsection{Ablation Study of Multi-instruction SFT Data}

To isolate the contribution of multi-task learning, we train multiple variants using Qwen2.5-7B-Instruct: the full model with all auxiliary tasks, three variants each removing one auxiliary task, and a single-task baseline trained only on event timeline summarization. All variants use identical hyperparameters. Table~\ref{tab:ablation_multitask} presents the results.

\begin{table}[t]
\centering
\caption{Multi-task learning ablation using Qwen2.5-7B-Instruct.}
\label{tab:ablation_multitask}
\begin{tabular}{l|ccc}
\toprule
\textbf{Configuration} & \textbf{F1↑} & \textbf{AR-1↑} & \textbf{$\tau$↑} \\
\midrule
\textbf{Full Model (Multi-task SFT)} & \textbf{76.2} & \textbf{73.8} & \textbf{0.87} \\
\midrule
w/o Temporal ordering & 72.9 & 68.2 & 0.83 \\
w/o Causal judgment & 73.1 & 68.0 & 0.79 \\
w/o Timeline completion & 72.7 & 67.6 & 0.82 \\
Main Task Only & 68.9 & 69.1 & 0.75 \\
\bottomrule
\end{tabular}
\end{table}

Results confirm that multi-task learning substantially boosts all metrics. The removal of timeline completion leads to the largest drop in overall quality, with a -3.5 F1 and -6.2 AR-1 decrease, highlighting its paramount importance for narrative coherence and content fidelity. The removal of causal judgment results in the most significant degradation in temporal ordering ($\tau$: -0.08), suggesting that understanding causality is critical for determining the correct event sequence beyond mere chronology. Although the absolute F1 drop from removing temporal ordering is moderate (-3.3), it causes a substantial AR-1 decrease (-5.6), confirming its foundational role in structuring event alignment.

The single-task baseline lags behind the full model by 7.3 F1 and 4.7 AR-1 points, demonstrating that auxiliary tasks provide essential inductive biases not learnable from the main task alone. Notably, even the weakest ablation variant (w/o Timeline completion) still outperforms the single-task baseline in F1 (72.7 vs. 68.9), confirming that each auxiliary task offers transferable benefits beyond its specific objective. This validates our multi-task learning framework as a comprehensive approach for improving timeline summarization capability.

\subsubsection{Results of Event Concise Summarization}

\begin{table}[t]
\centering
\caption{Comparison of event concise summarization performance. R-1, R-2, R-L, and Len represent ROUGE-1, ROUGE-2, ROUGE-L, and Length Compliance respectively.}
\label{tab:summary_gen}
\begin{tabular}{l|cccc}
\toprule
\textbf{Method} & \textbf{R-1↑} & \textbf{R-2↑} & \textbf{R-L↑} & \textbf{Len↑} \\
\midrule
DeepSeek-R1-671B (zero-shot) & 66.8 & 53.9 & 60.2 & 76.8 \\
DeepSeek-R1-671B (few-shot) & 69.5 & 57.2 & 63.8 & 80.5 \\
\midrule
Qwen2.5-3B-Instruct (SFT) & 66.3 & 53.7 & 59.8 & 78.3 \\
Qwen2.5-7B-Instruct (SFT) & 68.1 & 55.4 & 61.5 & 83.5 \\
\midrule
Qwen2.5-3B-Instruct (SFT+RL) & 68.3 & 56.2 & 60.6 & 85.5 \\
Qwen2.5-7B-Instruct (SFT+RL) & \textbf{71.9} & \textbf{59.8} & \textbf{65.8} & \textbf{88.2} \\
\bottomrule
\end{tabular}
\end{table}

We evaluate event concise summarization on 5K held-out clusters, comparing model scales and training strategies (Table~\ref{tab:summary_gen}). Reinforcement learning consistently improves length compliance by 4–7 percentage points (e.g., from 83.5\% to 88.2\% for Qwen2.5-7B-Instruct) and boosts ROUGE scores (e.g., +3.8 R‑1, +4.4 R‑2, and +4.3 R‑L for the 7B model), showing the reward signal encourages concise, information-rich summaries rather than simple truncation.
The Qwen2.5-7B-Instruct model with SFT+RL achieves 65.8\% ROUGE‑L and 88.2\% length compliance, substantially outperforming both zero-shot (60.2\% R‑L, 76.8\% Len) and few-shot (63.8\% R‑L, 80.5\% Len) DeepSeek-R1-671B baselines. These results suggest that task-specific multi-task fine-tuning
can substantially narrow the gap between compact domain models
and large general-purpose LLMs.
The notable gain in length compliance (88.2\% vs. 80.5\%) highlights that few-shot prompting struggles to enforce strict character limits consistently.

For production, we select the Qwen2.5-7B-Instruct SFT+RL model, which achieves the best balance between quality (65.8\% ROUGE‑L) and length compliance (88.2\%). This model significantly outperforms the 671B few-shot baseline in both semantic quality and constraint satisfaction, demonstrating that specialized RL training enables compact models to excel in production applications where display space is limited and quality requirements are stringent.

\subsection{Online Experiments}
\begin{table}[t]
\centering
\caption{Online A/B test results on Baidu Search. All values represent the relative improvement of the experimental group over the baseline.}
\label{tab:online_ab}
\small
\setlength{\tabcolsep}{2pt}
\begin{tabular}{l|l|ccc}
\toprule
\textbf{Timeline Model} & \textbf{Event Summary} & \textbf{CTR↑} & \textbf{Dwell Time↑} & \textbf{Depth↑} \\
\midrule
Main task Model (A) & Raw Title & - & - & - \\
Multi-task Model (B) & RL Summary & +5.5\% & +4.6\% & +4.4\% \\
\bottomrule
\end{tabular}
\end{table}

To evaluate the practical effectiveness of the multi-task supervised fine-tuning framework, we conducted multiple rounds of online A/B testing on Baidu Search from July to September 2025 under a controlled traffic deployment setting, covering more than 100 trending event timelines and collecting millions of user interaction samples.
The experimental setup was as follows: Group A (the baseline) used a model fine-tuned solely on the main task, employing raw document titles as event summaries; Group B used a model fine-tuned with multi-task learning, paired with event summaries optimized via reinforcement learning. Users were randomly assigned to the two groups to ensure a fair comparison of model performance.

Table~\ref{tab:online_ab} summarizes the online A/B test results.
Compared with the single-task baseline, the proposed framework shows consistent improvements across key user engagement metrics,
including CTR (+5.5\%), user dwell time (+4.6\%), and exploration
depth (+4.4\%). All reported improvements are statistically significant ($p < 0.05$).

These results suggest that the combination of multi-task learning and RL-based event summaries contributes to improved user
engagement and browsing experience.

\section{Downstream Application: Intra-timeline Heat Prediction}
\label{sec:heat_prediction}
Beyond generating query-driven event timelines, we explore a downstream application: predicting the relative heat of individual events within a timeline. In Baidu Search’s timeline presentation, not all events attract equal levels of public attention—some pivotal moments naturally receive substantially higher user interest than background context. Accurately identifying these high-attention events can improve timeline readability and information accessibility by helping users quickly locate major developments within long and complex event sequences. For example, timelines may optionally provide lightweight visual cues (e.g., subtle highlighting or expanded previews) to assist rapid navigation of important events, particularly in breaking-news scenarios where users seek concise situational awareness.

The proposed heat prediction mechanism is a non-personalized auxiliary signal rather than an engagement-optimization system. Heat labels are derived from aggregate historical page-view statistics across the overall user population, rather than user dwell time, post-click consumption behavior, or engagement-driven objectives. All users observe the same heat annotations for a given timeline, and the predicted heat levels do not alter event ranking or suppress low-heat events; all timeline events remain fully accessible.

A key challenge is that events within a timeline often occur in rapid succession with short temporal intervals, meaning that their page-view metrics have not yet stabilized when the timeline is first presented to users. Waiting for sufficient engagement signals to accumulate would delay potentially useful presentation adjustments, especially for emerging or fast-evolving news topics. Heat prediction addresses this cold-start setting by forecasting relative attention patterns directly from timeline structure and event semantics.

\subsection{Task Formulation and Training}

Given an event timeline $\mathcal{T} = \langle e_1, e_2, \ldots, e_n \rangle$ summarized by QDET, the heat prediction task aims to predict a relative heat level for each event:
\begin{equation}
f_{\text{heat}}: \mathcal{T} \rightarrow \mathbf{h} = [h_1, \ldots, h_n], \quad h_i \in \{\text{H1}, \text{H2}, \text{H3}, \text{H4}, \text{H5}\}
\end{equation}
where $h_i$ denotes the relative heat level of event $e_i$ within its timeline context. We adopt a quintile-based formulation: heat levels are assigned based on each event's percentile within the page view distribution of its timeline, ensuring cross-domain robustness by normalizing engagement differences across topics. We evaluate prediction quality using Accuracy, Macro-F1, and Mean Absolute Error (MAE).

We fine-tune the timeline summarization checkpoint on 3,000 heat-labeled timelines (2,400 training, 300 validation, 300 test), where heat labels are derived from page view quintiles aggregated over 7 days after publication. The model takes as input a query-driven event timeline produced by QDET and predicts heat levels for all constituent events, leveraging the timeline understanding acquired during timeline summarization training to model relative attention patterns within
event timelines.

\subsection{Experimental Results}

We compare three approaches: (1) QDET + Heat (Ours): our full model initialized with the timeline summarization checkpoint and fine-tuned on heat data; (2) From Scratch: training the same architecture (Qwen2.5-7B-Instruct) directly on heat data without timeline summarization to training; (3) DeepSeek-R1-671B: zero-shot prompting as a reference for general-purpose large language models.

\begin{table}[h]
\centering
\caption{Intra-timeline heat prediction performance on 300 test cases.}
\label{tab:heat_prediction}
\small
\begin{tabular}{l|ccc}
\toprule
\textbf{Method} & \textbf{Acc↑} & \textbf{Macro-F1↑} & \textbf{MAE↓} \\
\midrule
From Scratch & 40.5 & 36.8 & 1.15 \\
DeepSeek-R1-671B (zero-shot) & 49.2 & 45.8 & 0.88 \\
\textbf{QDET + Heat (Ours)} & \textbf{57.5} & \textbf{53.6} & \textbf{0.75} \\
\bottomrule
\end{tabular}
\end{table}
Table~\ref{tab:heat_prediction} shows that our method achieves 57.5\% accuracy, substantially outperforming comparison baselines. Training from scratch yields only 40.5\% accuracy, suggesting that timeline summarization initialization provides transferable timeline representations for modeling relative attention patterns within timelines. DeepSeek-R1-671B achieves 49.2\% accuracy in the zero-shot setting, while our specialized model surpasses it by 8.3 percentage points. Macro-F1 scores exhibit a similar trend (53.6\% vs. 36.8\% and 45.8\%, respectively). In particular, the from-scratch baseline struggles on extreme heat categories, indicating limited ability to capture relative importance relationships under low-resource supervision.
These results suggest that QDET’s structured timeline representations encode transferable representations useful for relative event attention estimation.

Given the relatively small scale of the heat prediction dataset, we further conduct 5-fold cross-validation to assess result stability. Our method achieves an average accuracy of 56.8\% ± 2.1\% across different splits, indicating reasonably consistent performance under limited-data conditions.
We acknowledge that the current evaluation remains limited in scale and is conducted entirely offline. Online evaluation of heat prediction presents additional challenges because visual highlighting itself directly influences user attention and click behavior, making it difficult to directly attribute behavioral changes to prediction quality through straightforward A/B testing.


\section{Limitations}
\label{sec:limits}
Although QDET demonstrates strong performance on query-driven timeline generation, we observe several important limitations in practical deployment scenarios.

One representative failure case involves long-running events whose key developments span substantially longer periods than the default retrieval window. In our current production system, candidate event retrieval primarily relies on a 30-day sliding temporal window to balance storage cost, retrieval latency, and online serving efficiency. While effective for most fast-evolving hot news events, this design can truncate important long-range event dependencies. For example, in the Yuhuaying child trafficking case, the first trial, second trial, and final judgment occurred across a time span significantly exceeding 30 days, causing temporally distant but semantically important events to be omitted from the same timeline. In such cases, QDET may generate incomplete timelines that capture local event bursts but fail to reconstruct the full long-term progression of the story.

In addition, QDET is primarily designed for high-attention news events with relatively dense reporting coverage, and its performance may degrade for long-tail queries or sparsely reported topics. The downstream heat prediction module also has limitations: predicted heat reflects aggregate public attention rather than factual importance or societal value. To reduce the risk of engagement-driven amplification, the current system intentionally restricts heat prediction to a non-personalized auxiliary signal that does not alter event ranking or suppress low-heat events.

\section{Conclusion}
\label{sec:conclusion}
This paper introduces QDET, a query-driven event timeline summarization system deployed on Baidu Search, which demonstrates that domain-specific multi-task learning enables compact models to approach the performance of general-purpose models up to 100$\times$ their size. Through RL-based event concise summarization and three auxiliary tasks—temporal ordering, causal judgment, and timeline completion—our 7B-parameter model achieves performance comparable to that of DeepSeek-R1-671B. Furthermore, our exploration of intra-timeline heat prediction confirms effective knowledge transfer. These results indicate that well-structured and logically coherent event timelines can improve user engagement and browsing experience in real-world search scenarios. By bridging the performance-efficiency gap through specialized training, QDET enables practical deployment of advanced natural language processing capabilities in resource-constrained industrial settings, transforming broad event contexts into focused narratives that help users understand complex event developments.

\bibliographystyle{unsrt}
\bibliography{ref}

\end{document}